\documentclass{article}

\usepackage{arxiv}

\usepackage[utf8]{inputenc} 
\usepackage[T1]{fontenc}    
\usepackage{hyperref}       
\usepackage{url}            
\usepackage{booktabs}       
\usepackage{amsfonts}       
\usepackage{nicefrac}       
\usepackage{microtype}      
\usepackage{lipsum}		
\usepackage{multirow}
\usepackage{multicol}
\usepackage{amsmath}
\usepackage{svg}
\usepackage{graphicx}
\usepackage[numbers]{natbib}
\usepackage{doi}
\usepackage{adjustbox}
\usepackage{siunitx}
\usepackage{textcomp}
\usepackage{longtable}
\usepackage{booktabs}

\title{Community-Based Model Sharing and Generalisation: Anomaly Detection in IoT Temperature Sensor Networks}

\author{ \href{https://orcid.org/0000-0002-2524-6299}{\includegraphics[scale=0.06]{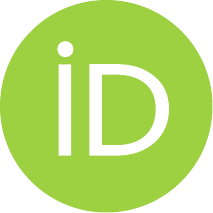}\hspace{1mm}Sahibzada Saadoon Hammad}\\
	Institute of New Imaging Technologies\\
	Universitat Jaume I\\
	Avd. Sos Baynat, s/n, Castelló de la Plana 12071 \\
	\texttt{hammad@uji.es} \\
	\And
    \href{https://orcid.org/0000-0002-8625-441X}{\includegraphics[scale=0.06]{orcid.pdf}\hspace{1mm}Joaquín Huerta Guijarro} \\
	Institute of New Imaging Technologies\\
	Universitat Jaume I\\
	Avd. Sos Baynat, s/n, Castelló de la Plana 12071 \\
	\texttt{joaquin.huerta@uji.es} \\
    \And
    \href{https://orcid.org/0000-0003-2540-4741}{\includegraphics[scale=0.06]{orcid.pdf}\hspace{1mm}Francisco Ramos}\\
	Institute of New Imaging Technologies\\
	Universitat Jaume I\\
	Avd. Sos Baynat, s/n, Castelló de la Plana 12071 \\
	\texttt{jromero@uji.es} \\
    \And
    \href{https://orcid.org/0000-0002-6431-1794}{\includegraphics[scale=0.06]{orcid.pdf}\hspace{1mm}Michael Gould Carlson} \\
	Institute of New Imaging Technologies\\
	Universitat Jaume I\\
	Avd. Sos Baynat, s/n, Castelló de la Plana 12071 \\
	\texttt{gould@uji.es} \\
    \And
    \href{https://orcid.org/0000-0002-9304-0719}{\includegraphics[scale=0.06]{orcid.pdf}\hspace{1mm}Sergio Trilles Oliver}\\
	Universitat Jaume I\\
	Avd. Sos Baynat, s/n, Castelló de la Plana 12071 \\
	\texttt{strilles@uji.es} \\
}

\renewcommand{\shorttitle}{Community-Based IoT Anomaly Detection}

\begin{document}
\maketitle

\begin{abstract}
The rapid deployment of Internet of Things (IoT) devices has led to large-scale sensor networks that monitor environmental and urban phenomena in real time. Communities of Interest (CoIs) provide a promising paradigm for organising heterogeneous IoT sensor networks by grouping devices with similar operational and environmental characteristics. This work presents an anomaly detection framework based on the CoI paradigm by grouping sensors into communities using a fused similarity matrix that incorporates temporal correlations via Spearman coefficients, spatial proximity using Gaussian distance decay, and elevation similarities. For each community, representative stations based on the best silhouette are selected and three autoencoder architectures (BiLSTM, LSTM, and MLP) are trained using Bayesian hyperparameter optimization with expanding window cross-validation and tested on stations from the same cluster and the best representative stations of other clusters. The models are trained on normal temperature patterns of the data and anomalies are detected through reconstruction error analysis. Experimental results show a robust within-community performance across the evaluated configurations, while variations across communities are observed. Overall, the results support the applicability of community-based model sharing in reducing computational overhead and to analyse model generalisability across IoT sensor networks.
\end{abstract}

\keywords{Communities of Interest \and Internet of Things \and Anomaly Detection \and Time-Series Data \and Deep Learning}

\section{Introduction}
The Internet of Things (IoT) has experienced remarkable growth, with billions of devices deployed across diverse environments to monitor various phenomena and collect data that support informed decision making \cite{lu2018internet, Granell2020}. These interconnected devices, organised within sensor networks, generate large volumes of continuous data, requiring effective data analysis, information extraction, and decision-making based on that information. These devices are susceptible to errors and malfunctions, creating streams of anomalous data. In critical domains such as environmental monitoring, smart cities, and industrial systems, it is essential to address these issues through robust anomaly detection techniques \cite{al2020survey} \cite{chatterjee2022iot}. Developing a unified anomaly detection system remains challenging due to the heterogeneity of IoT sensor deployments, variations in sensing capabilities, and the diverse geographical and contextual conditions under which data are collected \cite{diro2021comprehensive}. 

Conventional IoT anomaly detection applications have several limitations, such as limited scalability, high communication costs, privacy constraints, and a reduced ability to exploit contextual knowledge of the deployment environment \cite{zhang2024privacy}. Furthermore, developing universal anomaly detection models that perform consistently across heterogeneous sensor networks remains challenging due to diverse operational, environmental, and contextual factors \cite{balaji2023iot}. These challenges motivate the need for adaptive approaches capable of handling the non-stationarity, heterogeneity, and diversity inherent in IoT data.

In real-world IoT deployments, it is typically infeasible to train and maintain a separate anomaly detection model for each device. Many devices have limited historical data, limited computational capabilities, and no local expertise to implement or maintain complex models. This motivates the need for model-sharing approaches, where sensors with similar characteristics can leverage a common community-specific model, reducing computational costs and improving generalisation across the network.

Communities of Interest (CoI) have emerged as a promising paradigm for addressing heterogeneity in IoT sensor networks \cite{aldelaimi2020building}.  The Social IoT (SIoT) paradigm focuses on grouping and clustering devices with similar characteristics, such as local context, spatial location, and the operational environment \cite{roopa2019social}. This clustering in SIoT can be seen as a way of forming CoIs, where devices with similar properties naturally form communities. Within CoIs, sensors can share data and knowledge while preserving data privacy. Similarly, CoIs enable model sharing among sensors with comparable characteristics, facilitating more efficient and improved model generalisation by reducing the overall number of models required. 

This work presents a hierarchical clustering–driven Deep Learning (DL) anomaly detection framework for IoT devices applied to a temperature sensor network. The proposed approach organises devices into CoIs enabling model sharing within each community. Inter-device similarity is determined using multiple criteria, including temporal correlations in sensor measurements, spatial proximity, and elevation similarity. For anomaly detection, community-specific models are developed using three autoencoder architectures, namely Multi-Layer Perceptron (MLP), Long Short-Term Memory (LSTM), and Bidirectional Long Short-Term Memory Autoencoders (BiLSTM). This design enables the framework to capture diverse behavioural patterns while maintaining computational efficiency and adaptability across sensor communities.

The key contributions of our work are summarised below:

\begin{itemize}
    \item Formation of CoI through clustered multi-dimensional correlation analysis, incorporating temporal pattern and geospatial context (location and elevation).
    \item Development of community-specific anomaly detection models, trained using representative station selected from each community.
    \item Evaluation of model generalisability through a cluster-based training strategy, in which a single model is trained per community and validated using real-world temperature sensor data.
\end{itemize}

The remainder of this paper is organised as follows. Section \ref{sec:Related_works} presents a comprehensive review of related work on IoT anomaly detection and CoI. Section \ref{sec:methodology}  describes the methodology, detailing the individual stages, including exploratory data analysis, computation of correlation matrices, and sensor clustering. Section \ref{sec:algo} describes the considered autoencoder architectures. Section \ref{sec:exp_res} presents the experimental setup, while Section \ref{sec:res} details the results of the experiments. Section \ref{sec:conclusion} summarises the main findings and outlines directions for future research.

\section{Related Work}
\label{sec:Related_works}
\subsection{Anomaly Detection}

Anomaly detection in IoT has been applied across a wide range of domains, including network intrusion detection, fault detection, and industrial systems and machinery monitoring\cite{9047684}. Over time, anomaly detection approaches have evolved from manual detection and traditional statistical techniques to advanced Machine Learning (ML) and DL methods \cite{braei2020anomaly}. Furthermore, the rapid growth in the number of IoT devices and the massive volume of data they generate have driven the development of robust algorithms capable of capturing complex temporal patterns in IoT time-series data, thereby improving the effectiveness of anomaly detection \cite{rafique2024machine}.

Traditional anomaly detection methods, such as one-class Support Vector Machines (SVM) and isolation forests, have been applied in many studies; however, these approaches often struggle to capture temporal dependencies and to cope with the high dimensionality of IoT data \cite{li2019mad}. Similarly, statistical approaches exhibit limitations when addressing contextual anomalies, where deviations depend on surrounding conditions rather than absolute values. The authors in \cite{zahoor2025robust} report that, despite the effectiveness of one-class SVM and isolation forest algorithms in controlled experiments, these methods suffer from scalability and latency issues when handling high-frequency, high-dimensional sensor data, thereby constraining their applicability in resource-constrained, real-time IoT environments. Supervised learning approaches rely on datasets containing labelled instances of both normal behaviour and anomalies, which are often scarce, costly, and difficult to obtain in IoT settings, significantly limiting their adoption at scale \cite{diro2021comprehensive}. In contrast, unsupervised methods do not require labelled anomalies and instead learn patterns directly from the intrinsic structure of the data, making them particularly suitable for large-scale and dynamic IoT environments.

In recent years, DL algorithms have gained significant popularity in anomaly detection due to their ability to automatically learn hierarchical feature representations from raw data, thereby reducing the need for manual feature engineering. Various DL architectures such as Convolutional Neural Networks (CNNs), Recurrent Neural Networks (RNNs), Generative Adversarial Networks (GANs), and autoencoders have been used for anomaly detection \cite{li2019mad, He2019TemporalCN, du2021gan, malhotra2016lstm}. The authors in \cite{zamanzadeh2024deep} provide a comprehensive survey of DL–based anomaly detection methods, categorising existing approaches into forecasting- and reconstruction-based paradigms. Reconstruction‑based methods learn latent representations of normal behaviour and identify anomalies by detecting large reconstruction errors. Numerous autoencoders variants have been explored in the literature, including LSTM autoencoders, CNN autoencoders, and Variational Autonencoders (VAE). The authors in \cite{chen2020unsupervised} propose a Sliding Window based Convolutional Variational Autoencoder (SWCVAE) to detect real-time spatial and temporal anomalies in industrial robots. Hybrid approaches that combine attention mechanisms with autoencoders have been investigated to enhance anomaly detection by capturing both global temporal dependencies and local reconstruction errors in real-world time-series data from the Numenta Anomaly Benchmark (NAB) dataset \cite{najafi2024attention}. Another hybrid model integrating CNNs with recurrent autoencoders has been proposed for anomaly detection in IoT time-series, where CNN layers extract spatial features and recurrent units model temporal dependencies \cite{yin2020anomaly}. Anomaly detection in smart building environments has also been studied using a heterogeneous IoT dataset comprising 14 sensors. This work compared multiple approaches, including statistical techniques, clustering, one-class SVM, isolation forest, and DL models such as CNN and LSTM. DL–based approaches, particularly autoencoder variants, consistently achieved more robust detection performance across diverse anomaly scenarios when compared with classical ML baselines \cite{majib2023detecting}.

Finally, Graph Neural Networks (GNNs) have also been explored for anomaly detection in IoT and multi-sensor systems. The authors in \cite{li2022correlation} propose a correlation-based anomaly detection framework for multi-sensor systems, in which temporal correlation graphs are constructed and processed using a structure-aware GNN. More recently, GNN-based approaches have been applied to air quality time-series data, exploiting spatio-temporal correlations among data streams collected by distributed sensors \cite{LinGNN}.

\subsection{Communities of Interest in IoT}

The concept of CoI originates from the SIoT paradigm \cite{atzori2012social}. In SIoT, devices within the network are modelled as social entities capable of establishing autonomous relationships with other devices. Based on shared characteristics or properties, these devices form communities with common interests, referred to as CoIs \cite{bao2013scalable}. CoIs are defined as one of the key social trust parameters within trust evaluation models for SIoT \cite{sagar2020towards}. Several trust management systems for IoT have been proposed leveraging the CoI concept. The authors in \cite{djedjig2020trust} introduce a dynamic hierarchical trust model that exploits CoIs to adjust trust levels based on device mobility and shared community context. Similarly, \cite{abderrahim2017tmcoi} propose TMCoI-SIoT, a trust management system in which IoT devices are organised into communities. In this approach, devices are clustered, with each cluster representing a CoI that groups devices with similar interests. Furthermore, the study in \cite{aldelaimi2020building} proposes a Dynamic Community of Interest Model (DCIM), which allows IoT devices to dynamically form CoI based on shared interests and the emergence of new inter-device relationships.

While previous studies have primarily concentrated on security-related aspects, they have shown that organising devices into interest-based communities can improve scalability and robustness. Building on these findings, the proposed approach leverages community formation as a mechanism for sharing anomaly detection models among IoT devices with similar operational and environmental characteristics. Sensors are clustered using multiple similarity criteria, including temporal correlations, spatial proximity, and elevation similarity, enabling the deployment of community-shared anomaly detection models. For each community, autoencoder-based models (e.g., MLP-AE, LSTM-AE, and BiLSTM-AE) are trained using representative stations and subsequently shared within the corresponding cluster. This strategy reduces the number of models required while preserving the ability to detect anomalous behaviour across groups of similar sensors.

\section{Methodology}
\label{sec:methodology}

\begin{figure}
    \centering
    \includegraphics[  width=\textwidth,  height=0.71\textheight,  keepaspectratio]{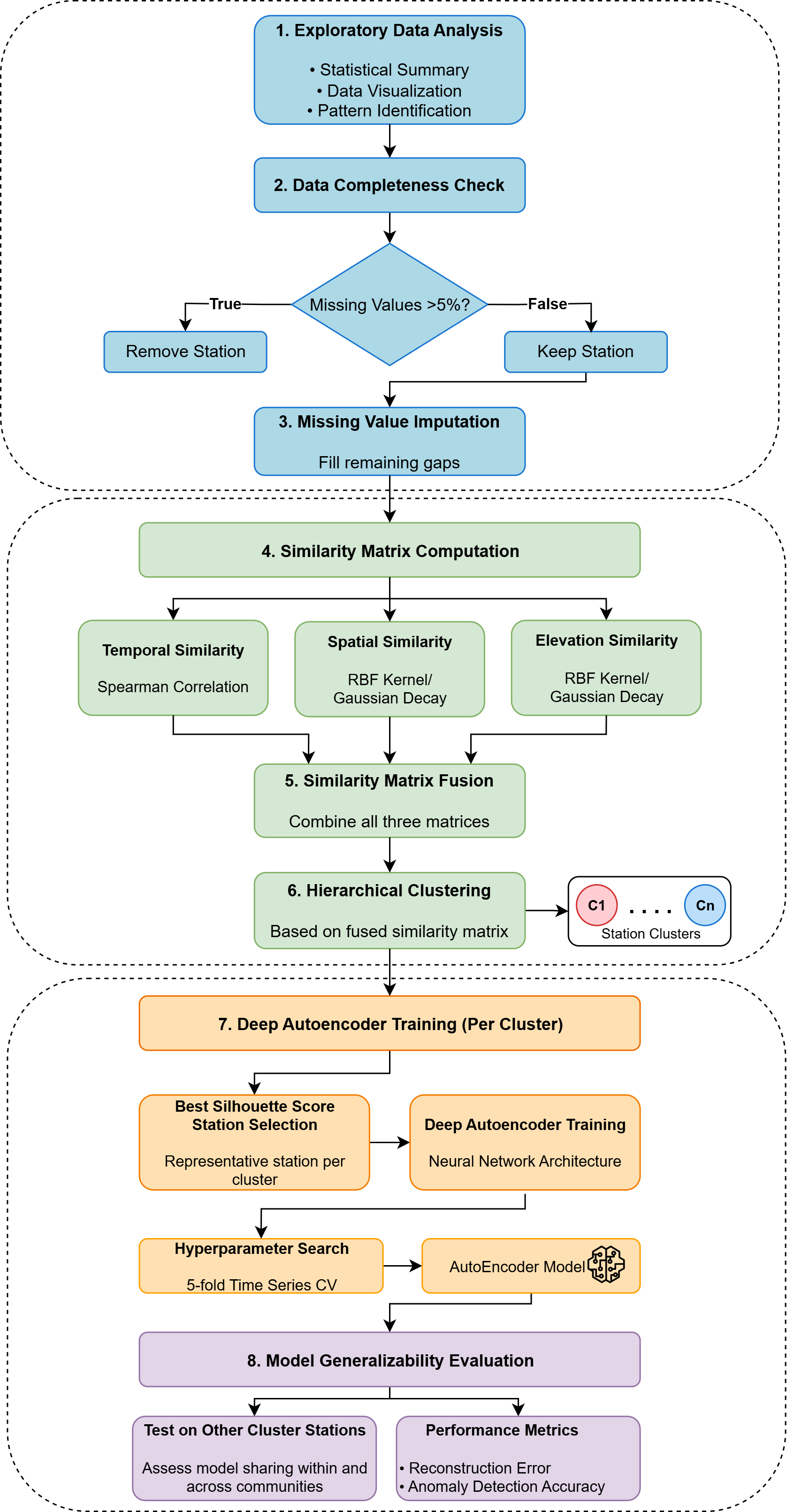}
    \caption{Schematic overview of community-based anomaly detection framework}
    \label{fig:methodology}
\end{figure}

This study proposes a systematic framework for community-driven anomaly detection in IoT temperature sensor data collected from meteorological stations. The methodology combines structured data preprocessing, multi-dimensional similarity–based clustering, and deep autoencoder models to identify abnormal sensor behaviour patterns. The complete workflow is illustrated in Figure \ref{fig:methodology}, and starts with exploratory data analysis and quality assessment, followed by hierarchical clustering of stations based on temporal, spatial proximity, and elevation similarity. For each cluster, representative stations are selected to train community-specific autoencoder models, specifically MLP, LSTM and BiLSTM architectures using Bayesian hyperparameter optimisation with expanding window cross-validation. 

The trained models learn to reconstruct normal temperature patterns, enabling anomaly detection through analysis of reconstruction error analysis. Finally, the generalisability of cluster-specific models is assessed by evaluating them on stations within the same cluster, verifying whether the trained models apply effectively across sensors with similar characteristics. The models are also evaluated on stations belonging to other clusters to analyse their ability to generalise beyond the communities used for training.

\subsection{Exploratory Data Analysis and Data Preparation}
\label{sec:EDA}

\begin{figure}
    \centering
    \includegraphics[width=0.8\textwidth,  height=0.8\textheight,  keepaspectratio]{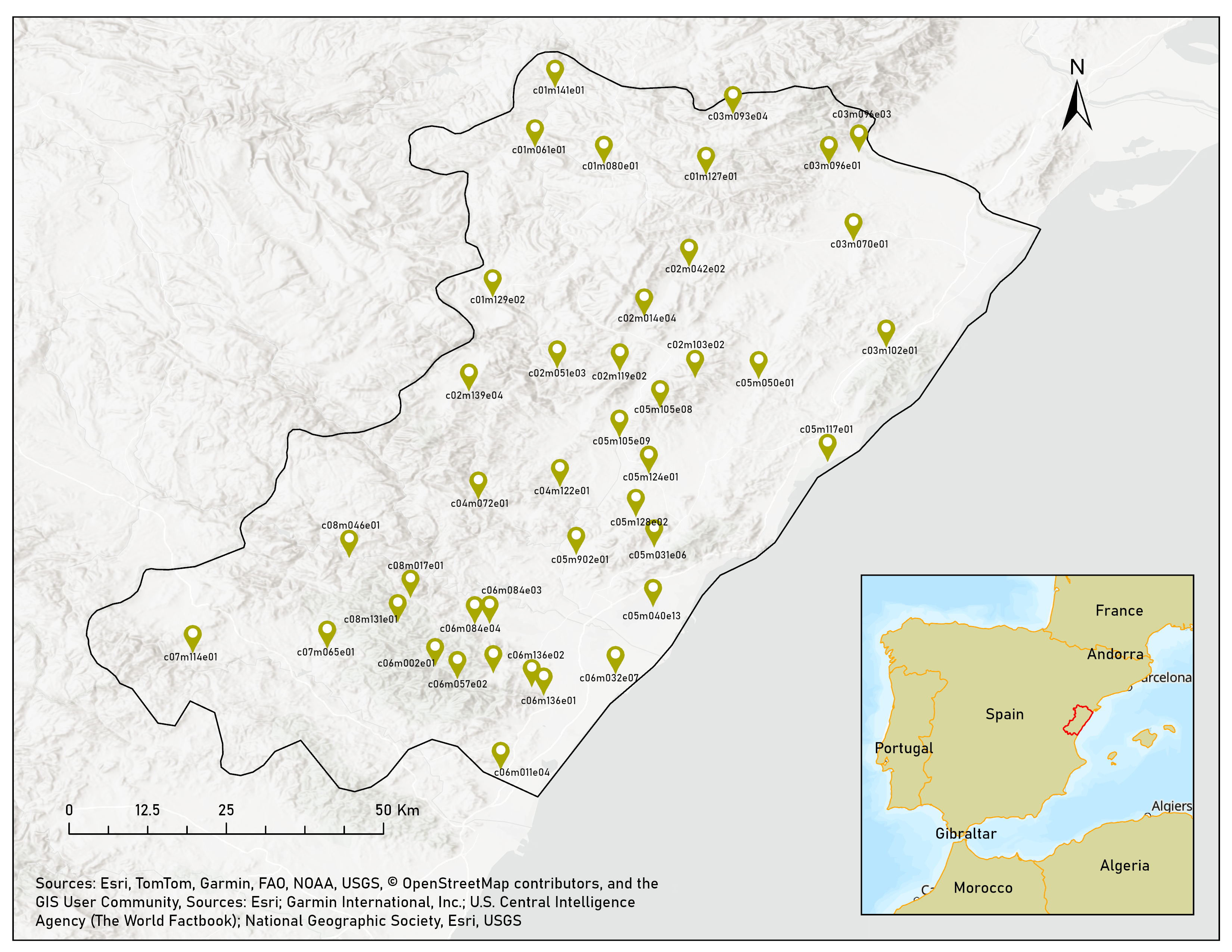}
    \caption{Location of meteorological stations in the sensor network.}
    \label{fig:stations_map}
\end{figure}

The dataset is collected from a network of meteorological sensors distributed across the Valencian Community (Spain), comprising more than 600 stations currently deployed, and operated by the Associació Valenciana de Meteorologia “Josep Peinado” (AVAMET) \cite{avamet_meteoxarxa}. For the purposes of this study, only the subset of 41 stations located in the province of Castelló is considered. Measurements are recorded at a fixed sampling interval of ten minutes, yielding 144 observations per day for each station. The geographical spread of the sensors is shown on the map in Figure \ref{fig:stations_map}.

\begin{figure}
    \centering
    \includegraphics[  width=0.7\textwidth,  height=0.7\textheight,  keepaspectratio]{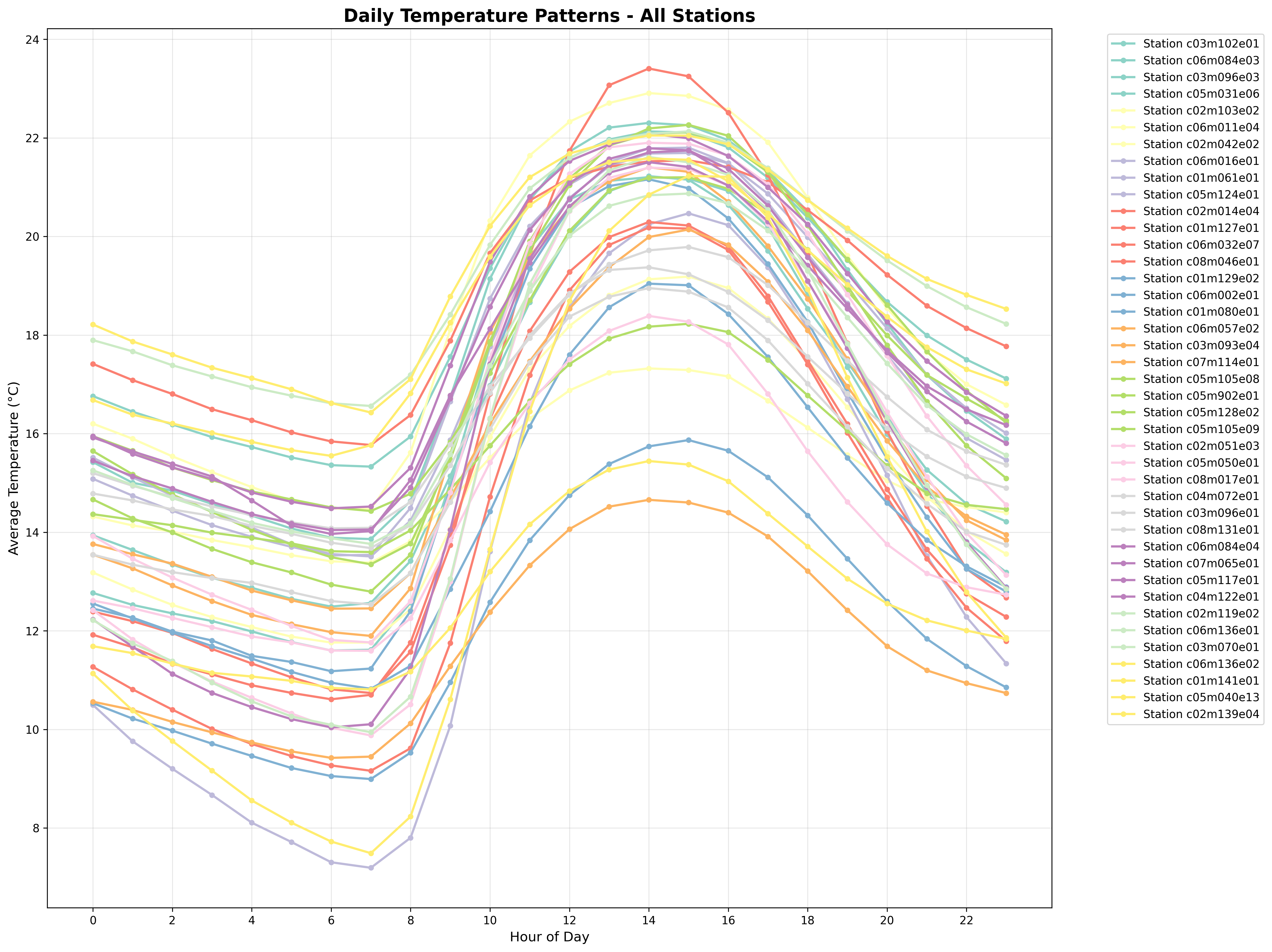}

    \caption{Average daily temperature patterns across the selected meteorological stations.}
    \label{fig:avg_temp}
\end{figure}

The plot in Figure \ref{fig:avg_temp} illustrates the average daily temperature patterns across all stations, providing a foundation for the Exploratory Data Analysis (EDA). All stations show a clear diurnal cycle, with temperature peaking in the afternoon (around 2 PM) and reaching their lowest values in the early morning around (6 AM). The figure also highlights variations in temperature amplitude, with some stations showing pronounced day–night differences between their minimum and maximum temperature while others display a more limited thermal range.

\begin{figure}
    \centering
    \includegraphics[  width=0.7\textwidth,  height=0.70\textheight,  keepaspectratio]{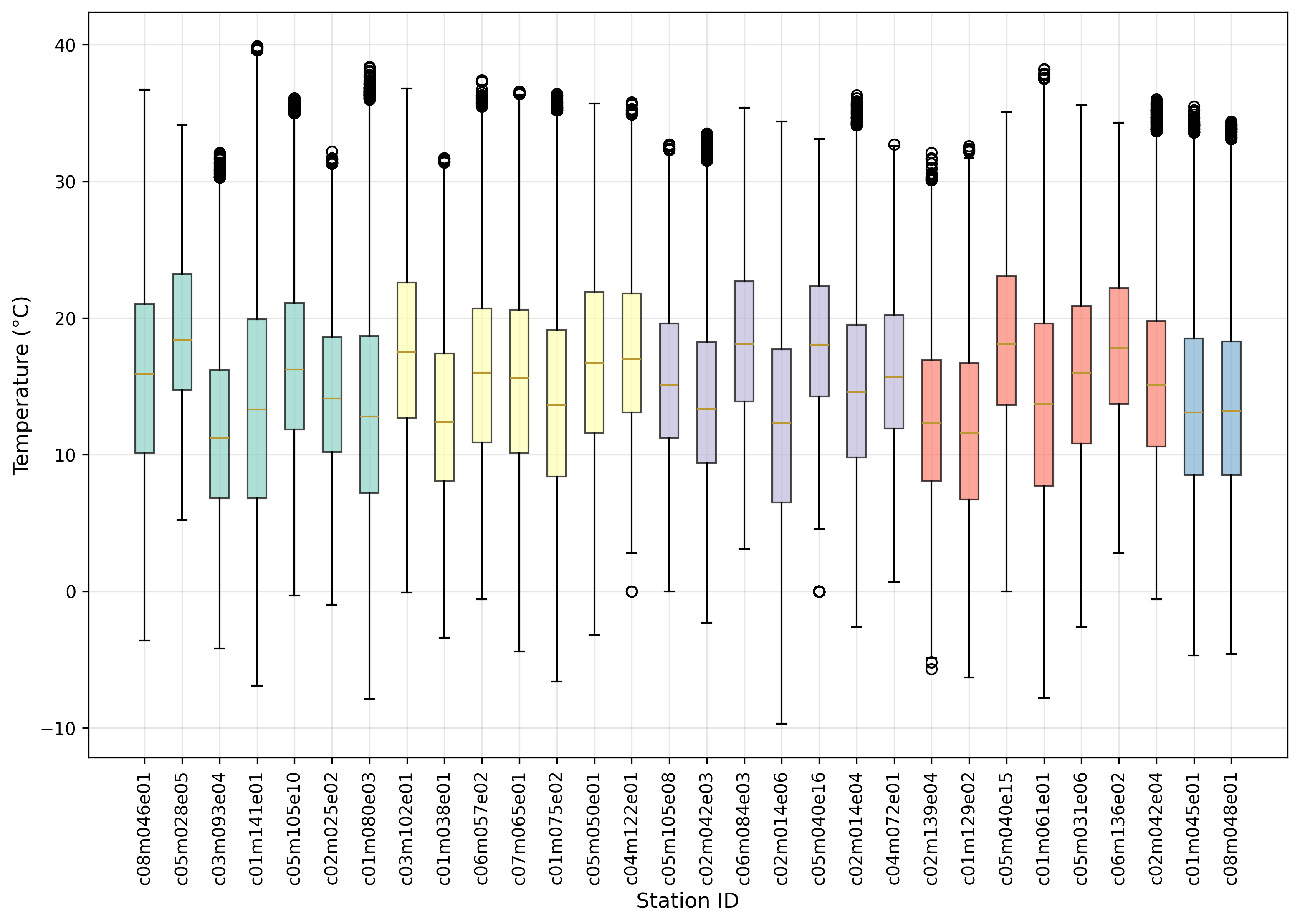}
    \caption{Distribution of daily temperature values across stations represented as boxplots.}
    \label{fig:dist_box}
\end{figure}

To further examine the data distribution, a boxplot chart is used to analyse the daily distribution of temperature time series data in Figure \ref{fig:dist_box}. The boxplots illustrate the central tendency, dispersion, and presence of extreme outliers. Most stations exhibit temperature ranges typically between \SI{0}{\degreeCelsius} to  \SI{35}{\degreeCelsius}, with median values generally lying between \SI{15}{\degreeCelsius} - \SI{20}{\degreeCelsius}. However, notable differences in interquartile ranges suggest higher variability at certain stations. Additionally, several stations present outliers exceeding \SI{40}{\degreeCelsius} and falling below \SI{-5}{\degreeCelsius}, which may be attributed to extreme weather conditions or sensor malfunctions.

\begin{figure}
    \centering
    \includegraphics[  width=0.7\textwidth,  height=0.7\textheight,  keepaspectratio]{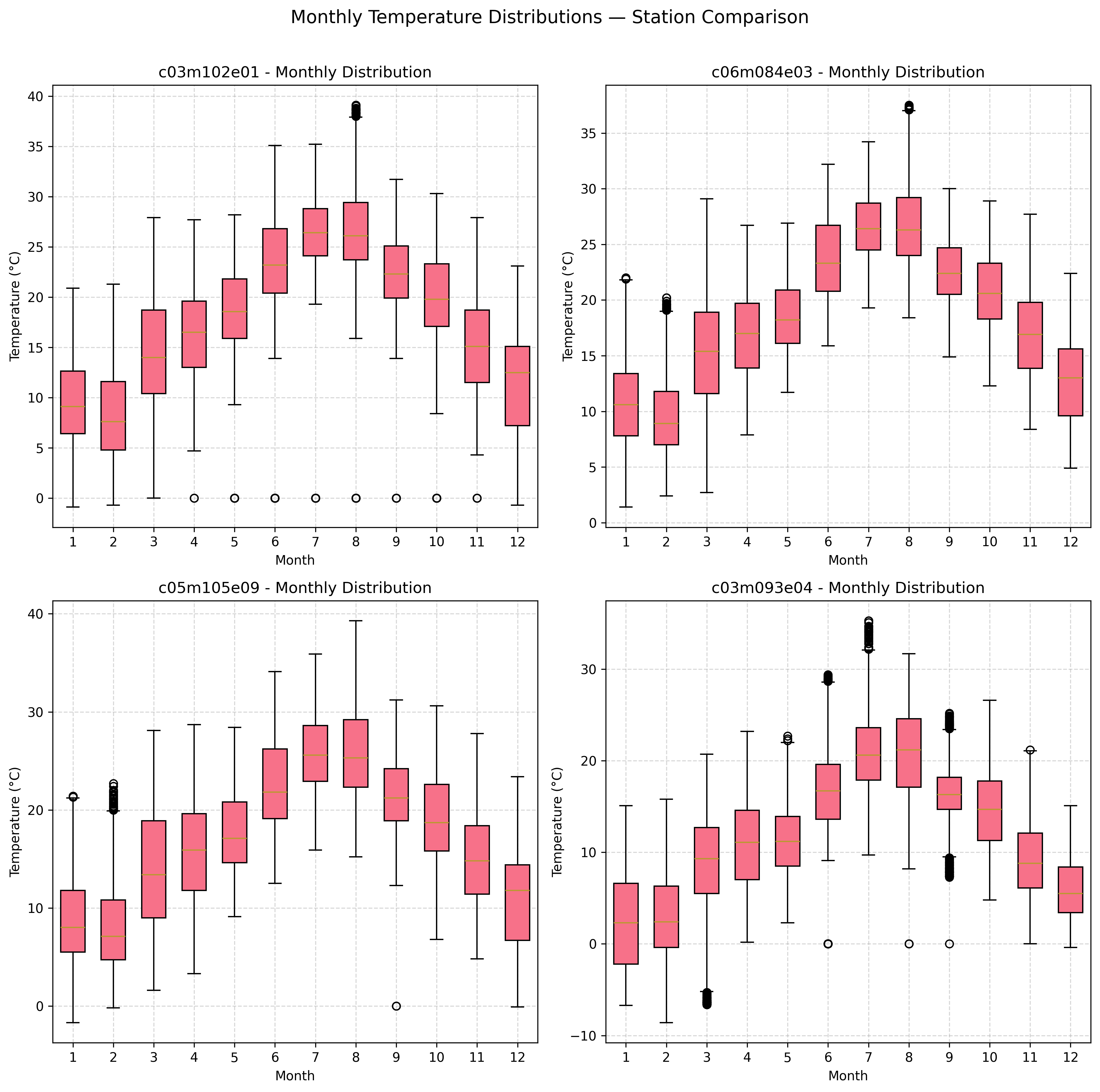}
    \caption{Monthly temperature distributions for a subset of representative stations.}
    \label{fig:month_dist}
\end{figure}

The monthly temperature distributions for four weather stations (\textit{c03m102e01}, \textit{c06m084e03}, \textit{c05m105e09}, and \textit{c03m093e04}) are visualised using box plots in Figure \ref{fig:month_dist}. The figure shows the comparative distinct seasonal temperature patterns across four meteorological stations. Station \textit{c03m102e01} exhibits the most pronounced seasonal variability, with winter months showing temperatures approaching \SI{0}{\degreeCelsius}, while summer temperatures reach values close to \SI{39}{\degreeCelsius}. Similarly, station \textit{c05m105e09} exhibits a comparable median temperature profile, with values peaking during July and August. Station \textit{c06m084e03} is characterised by milder winter temperatures, typically around \SI{5}{\degreeCelsius}, and relatively cooler summers conditions compared with \textit{c03m102e01}. Station \textit{c03m093e04} remains cooler throughout the year, with several months recording sub-zero temperatures and a broader temperature range during summer. Additionally, the presence of outliers during the summer months for stations \textit{c03m093e04} and \textit{c06m084e03} suggests the occurrence of occasional extreme temperature events.

\subsection{Similarity Matrix Computation}
We compute three independent similarity matrices to capture complementary relationships among sensors: temporal similarity, quantified using Spearman correlation coefficients; spatial similarity, derived from geographical distance; and elevation similarity, computed based on altitude differences. These matrices are subsequently integrated to obtain a single fused similarity matrix, which encapsulates the multi-dimensional relationships between sensors. 

\subsubsection{Temporal Correlation}

\begin{figure}
    \centering
    \includegraphics[width=0.7\textwidth,  height=0.7\textheight,  keepaspectratio]{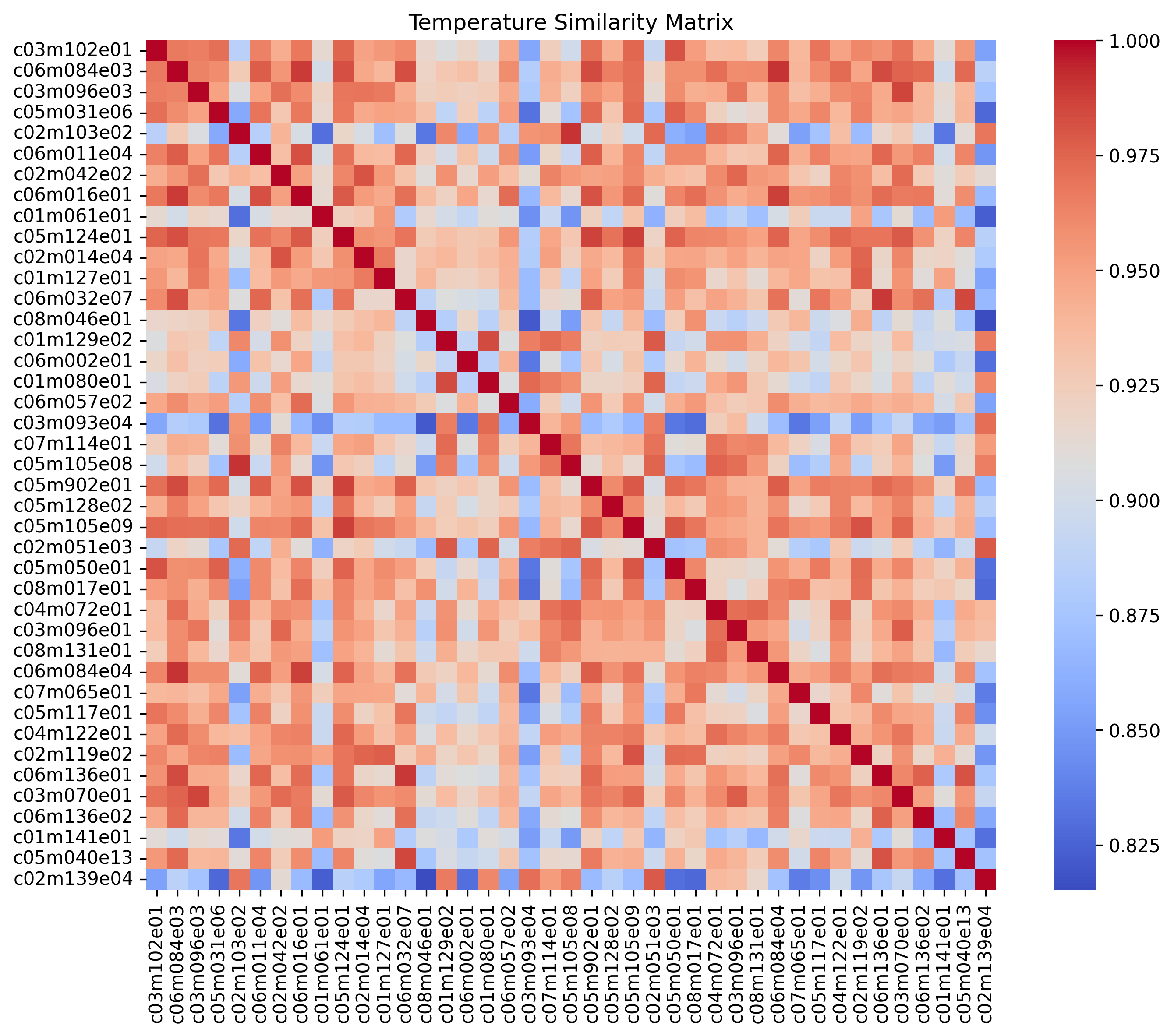}
    \caption{Temporal correlation matrix computed using Spearman method.}
    \label{fig:temporal_matrix2}
\end{figure}

The temporal correlation is computed using pairwise Spearman correlation coefficients, as shown in Figure \ref{fig:temporal_matrix2}. The values for the correlation co-efficient lie in the range $[-1, 1]$, where positive values shows a strong positive relationship, while negative values shows strong negative relationship. A value close to zero suggests little or no monotonic relationship between the variables. For two sensors $i$ and $j$ with temperature readings $x_i(t)$ and $x_j(t)$ observed over time $t = 1, 2, \dots, T$, the temporal correlation $r_{ij}$ is defined as follows in \eqref{eq:spearman_corr}:


\begin{equation}
r_{ij} =
\frac{\sum_{t=1}^T \left( R_i(t) - \bar{R}_i \right)
\left( R_j(t) - \bar{R}_j \right)}
{\sqrt{\sum_{t=1}^T \left( R_i(t) - \bar{R}_i \right)^2}
\sqrt{\sum_{t=1}^T \left( R_j(t) - \bar{R}_j \right)^2}}
\label{eq:spearman_corr}
\end{equation}

where $R_i(t) = \mathrm{rank}\!\left(x_i(t)\right)$ and
$R_j(t) = \mathrm{rank}\!\left(x_j(t)\right)$ denote the ranks of the
temperature measurements of sensors $i$ and $j$ at time $t$,
respectively, and $\bar{R}_i$ and $\bar{R}_j$ represent their
corresponding mean ranks.

\subsubsection{Spatial Correlation}

Spatial similarity is computed based on the pairwise Euclidean distances between the sensors. The geographical coordinates (latitude and longitude) of each sensor are used to calculate the distance $d_{ij}$ between each pair of sensors $i$ and $j$. A Gaussian distance decay function, defined in \eqref{eq:spatial_corr}, is then applied to map distances to similarity values, such that spatially closer sensors are assigned higher similarity scores. The parameter $\sigma$ in the Gaussian kernel is set as the standard deviation of all pairwise inter-sensor distances, enabling adaptive scaling on the spatial distribution of the network. 

\begin{equation}
s_{ij} = \exp \left( - \frac{d_{ij}^2}{2 \sigma^2} \right)
\label{eq:spatial_corr}
\end{equation}

\begin{figure}
    \centering
    \includegraphics[width=0.7\textwidth,  height=0.7\textheight,  keepaspectratio]{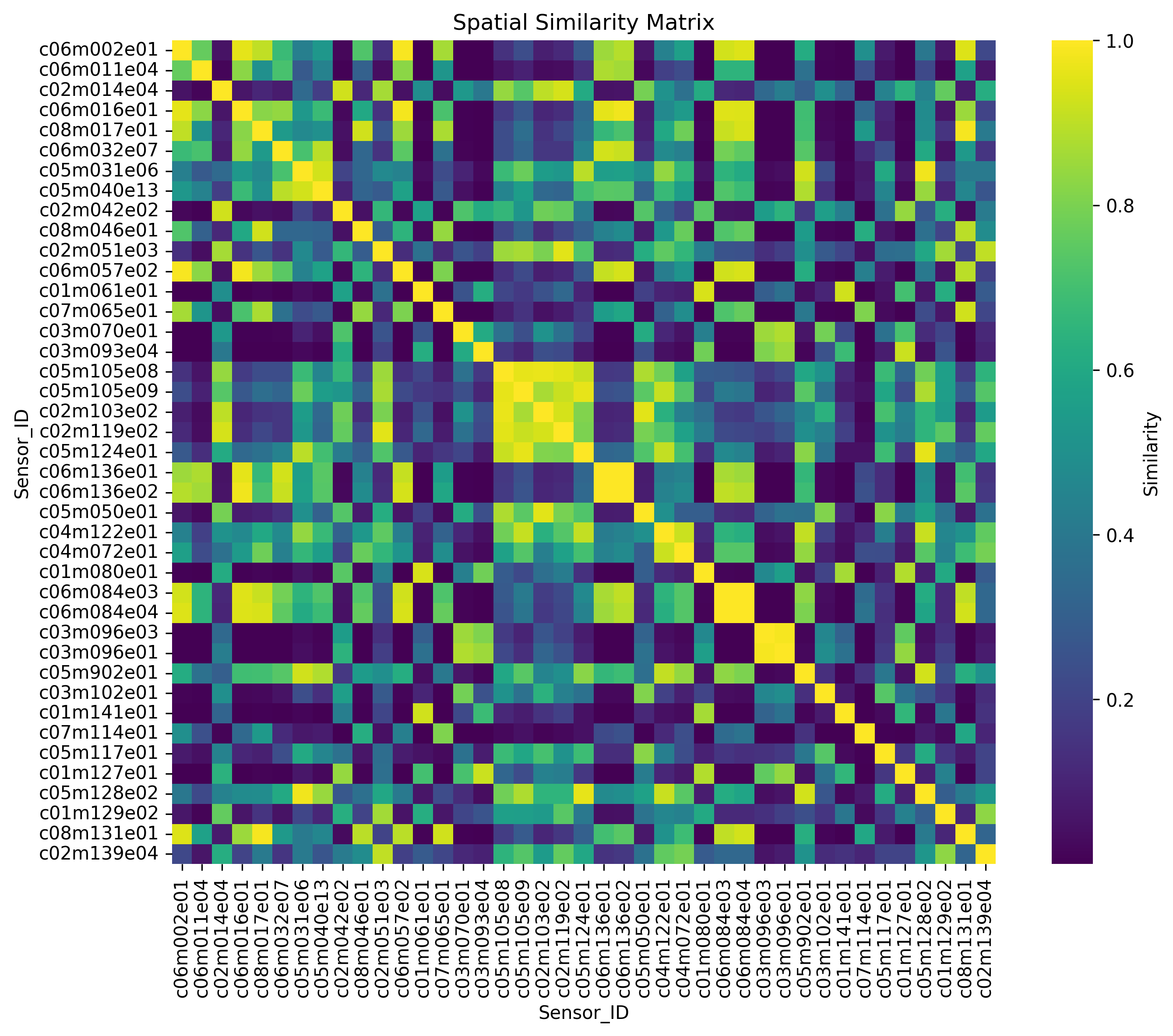}
    \caption{Spatial similarity matrix computed using a Gaussian distance-decay function.}
    \label{fig:spatial_matrix2}
\end{figure}

This formulation is consistent with prior studies in air quality monitoring and spatio-temporal graph modeling \cite{li2018diffusioncrn,LinGNN}. The resulting spatial correlation values lie in the range $(0, 1]$ shown in Figure \ref{fig:spatial_matrix2}.

\subsubsection{Elevation Correlation}

\begin{figure}
    \centering
    \includegraphics[width=0.7\textwidth,  height=0.7\textheight,  keepaspectratio]{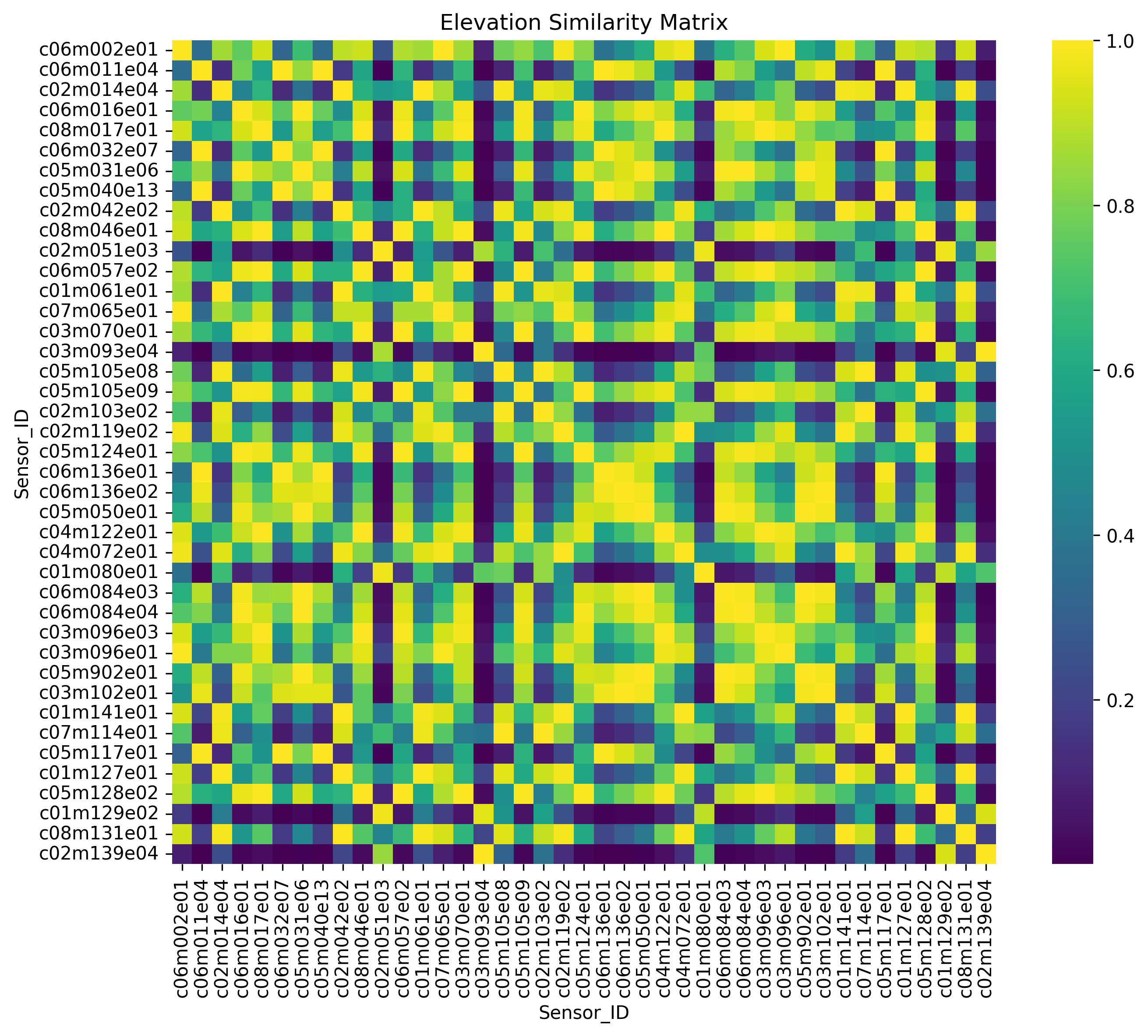}
    \caption{Elevation correlation matrix computed using a Gaussian distance-decay function.}
    \label{fig:elevation_matrix2}
\end{figure}

The elevation similarity is calculated on the absolute elevation differences between sensors. The elevation values $e_i$ and $e_j$ for each pair of sensors $i$ and $j$ are used to calculate the difference $d_{ij}^{(e)} = |e_i - e_j|$. A Gaussian distance-decay function, defined in  \eqref{eq:elevation_corr}, is then applied to map the elevation differences to similarity values, such that sensors with similar altitudes are assigned higher similarity scores. The parameter $\sigma$ in the Gaussian kernel is set as the standard deviation of all pairwise elevation differences, providing adaptive scaling based on the altitude distribution of the network. The resulting elevation similarity values lie in the range $(0, 1]$, as illustrated in Figure \ref{fig:elevation_matrix2}.

\begin{equation}
s_{ij}^{(e)} = \exp \left( - \frac{ \left( d_{ij}^{(e)} \right)^2 }{2 \sigma^2} \right)
\label{eq:elevation_corr}
\end{equation}

The resulting elevation similarity values lie in the range $(0, 1]$.

\subsection{Clustering}

Hierarchical clustering is employed to group stations based on the fused similarity matrix. This agglomerative clustering strategy iteratively merges the most similar stations or clusters, resulting in a dendrogram that captures the hierarchical relationships among all stations. Prior to clustering, the fused similarity matrix is transformed into a dissimilarity matrix, which is then used as input to the hierarchical clustering algorithm.

\begin{figure}
    \centering
    \includegraphics[width=0.9\textwidth,  height=0.9\textheight,  keepaspectratio]{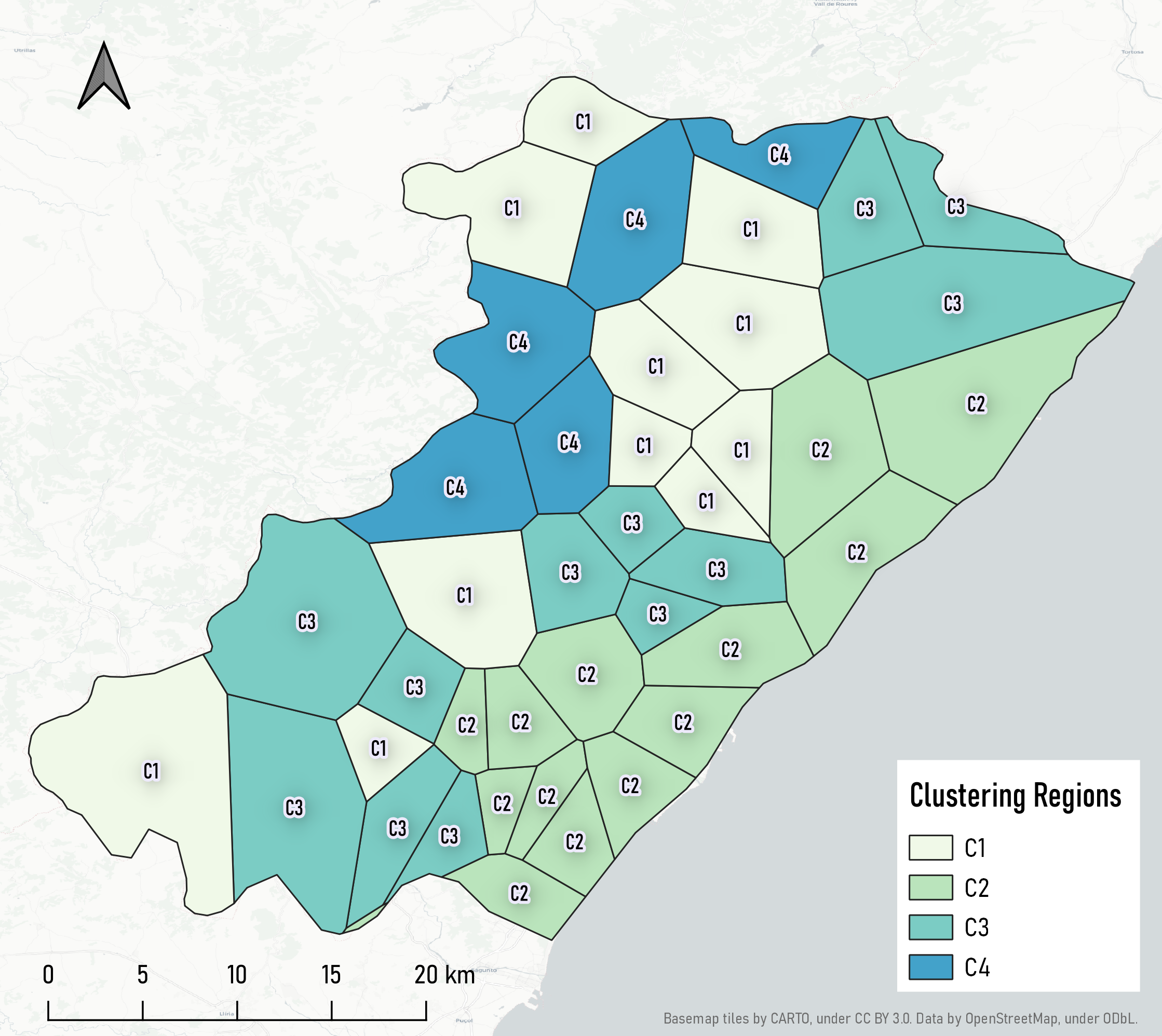}
    \caption{Voronoi polygons derived from the clustering of meteorological stations}
    \label{fig:cluster_regions}
\end{figure}

\begin{table*}[htbp]
\caption{Sensors belonging to Cluster~1, along with their silhouette scores and elevation values.}
\centering
\begin{tabular}{llcc}
\toprule
\textbf{Sensor ID} & \textbf{Municipality} &
\textbf{Silhouette Score} & \textbf{Elevation (m)} \\
\midrule
\textit{c02m042e02} & Catí                  & 0.9535 & 661 \\
\textit{c02m014e04} & Ares del Maestrat     & 0.9469 & 690 \\
\textit{c01m061e01} & Forcall               & 0.9283 & 692 \\
\textit{c08m131e01} & Villamalur            & 0.8511 & 637 \\
\textit{c01m127e01} & Vallibona             & 0.8479 & 646 \\
\textit{c05m105e08} & La Serra d'en Galceran& 0.8239 & 744 \\
\textit{c01m141e01} & Sorita                & 0.8132 & 620 \\
\textit{c07m114e01} & Toras                 & 0.7970 & 773 \\
\textit{c02m119e01} & La Torre d'en Besora  & 0.5944 & 582 \\
\textit{c02m103e02} & La Serratella         & 0.5933 & 781 \\
\textit{c04m072e01} & Llucena               & 0.5358 & 585 \\
\bottomrule
\end{tabular}
\label{tab:cluster1_sensors}
\end{table*}

The clustering performance is evaluated for different numbers of clusters, with \( K \in \{2, 3, 4, 5, 6, 7, 8\} \), using the silhouette score as the evaluation metric. Hierarchical clustering applied to the fused similarity matrix identifies four distinct clusters, with the optimal configuration selected by maximizing the silhouette score. Sensors with a silhouette score below 0.25 were excluded from the experiments to ensure that only stations exhibiting meaningful cluster structure contributed to the community-based model training and evaluation \cite{larose2015data}. The resulting spatial distribution of the clusters is shown in Figure \ref{fig:cluster_regions}.

Table \ref{tab:cluster1_sensors} summarises the stations belonging to Cluster~1 (C1) along with their corresponding silhouette scores, which range from 0.5358 to 0.9535 across stations located at different elevations, with La Serratella (\textit{c02m103e02}) being the highest-elevation station at 781 meters. The station \textit{c02m042e02} (Catí) achieved the highest silhouette score of 0.9535 at an elevation of 661 meters, indicating that it is the most representative station of this cluster. Consequently, this station is selected as the reference for training the anomaly detection models associated with C1.

Cluster~2 (C2), summarised in Table \ref{tab:cluster2_sensors}, exhibits a distinct profile compared to Cluster~1. It comprises 13 stations distributed across multiple municipalities, including Castelló de la Plana, Onda, and Artana. The silhouette scores within this cluster display considerable variability, ranging from 0.1421 to 0.9473, with two stations (\textit{c06m084e04} and \textit{c06m016e01}) exhibiting values below 0.20, indicating weak cluster cohesion and potential ambiguity in their assignment. Most stations in this cluster are located at elevations below 200 meters, indicating a predominantly low-altitude deployment.

\begin{table}[htbp]
\caption{Sensors belonging to Cluster~2, along with their silhouette scores and elevation values.}
\centering
\begin{tabular}{llcc}
\toprule
\textbf{Sensor ID} & \textbf{Municipality} & \textbf{Silhouette Score} & \textbf{Elevation (m)} \\
\midrule
\textit{c06m136e02} & La Vilavella              & 0.9473 & 115 \\
\textit{c05m040e13} & Castello de la Plana      & 0.9415 &  22 \\
\textit{c06m136e01} & La Vilavella              & 0.9380 &  50 \\
\textit{c06m032e07} & Borriana                  & 0.9362 &  12 \\
\textit{c06m011e04} & Almenara                  & 0.9299 &  32 \\
\textit{c03m102e01} & Santa Magdalena de Polpis & 0.9123 & 123 \\
\textit{c05m117e01} & Torreblanca               & 0.8942 &   0 \\
\textit{c05m902e01} & Sant Joan de Moro         & 0.8798 & 184 \\
\textit{c05m050e01} & Les Coves de Vinroma      & 0.7908 & 186 \\
\textit{c06m084e03} & Onda                      & 0.7702 & 196 \\
\textit{c05m031e06} & Borriol                   & 0.7648 & 223 \\
\textit{c06m084e04} & Onda                      & 0.1436 & 248 \\
\textit{c06m016e01} & Artana                    & 0.1421 & 262 \\
\bottomrule
\end{tabular}
\label{tab:cluster2_sensors}
\end{table}

Cluster~3 (C3), summarised in Table \ref{tab:cluster3_sensors}, consists of 12 stations distributed across several municipalities, including Cirat, Rossel and Ain, with silhouette scores ranging from 0.1395 to 0.9114. Elevation values span a moderate range between 305 meters to 517 meters, with most stations concentrated between 330 and 401 meters. The station \textit{c08m046e01} (Cirat), situated at 374 meters and achieving a silhouette score of 0.9414, is selected as the representative for training purposes the anomaly detection models associated with this cluster.

\begin{table}[htbp]
\caption{Sensors belonging to Cluster~3, along with their silhouette scores and elevation values.}
\centering
\begin{tabular}{llccc}
\toprule
\textbf{Sensor ID} & \textbf{Municipality}  &
\textbf{Silhouette Score} & \textbf{Elevation (m)} \\
\midrule
\textit{c08m046e01}  & Cirat                   & 0.9414 & 374 \\
\textit{c08m017e01} & Ayodar                  & 0.9295 & 380 \\
\textbf{c03m096e03} & Rossell                & 0.9197 & 392 \\
\textit{c04m122e01} & Les Useres             & 0.9114 & 401 \\
\textit{c06m057e02} & Eslida                  & 0.8399 & 342 \\
\textit{c05m128e02} & Vilafamés              & 0.7027 & 350 \\
\textit{c03m070e01} & la Jana                 & 0.6843 & 330 \\
\textit{c06m002e01} & Ain                     & 0.6273 & 510 \\
\textit{c07m065e01} & Gaibiel                & 0.5687 & 517 \\
\textit{c05m105e09} & La Serra d'en Galceran  & 0.5489 & 314 \\
\textit{c03m096e01} & Rossell                & 0.5425 & 475 \\
\textit{c05m124e01} & La Vall d'Alba          & 0.1395 & 305 \\
\bottomrule
\end{tabular}
\label{tab:cluster3_sensors}
\end{table}

Table \ref{tab:cluster4_sensors} summarises Cluster~4 (C4), which comprises five stations with silhouette scores ranging from 0.7663 to 0.9708. This cluster is geographically concentrated in high-elevation mountainous municipalities, including Vilafranca, La Pobla de Benifassà, Vistabella del Maestrat, Culla, and Morella. The silhouette scores indicate strong intra-cluster cohesion, particularly for Clusters~1 and~4, while the elevation data suggest that geographical and altitudinal proximity play a significant role in the clustering outcome.

\begin{table}[htbp]
\caption{Sensors belonging to Cluster~4, along with their silhouette scores and elevation values.}
\centering
\begin{tabular}{llcc}
\toprule
\textbf{Sensor ID} & \textbf{Municipality} & \textbf{Silhouette Score} & \textbf{Elevation (m)} \\
\midrule
\textit{c01m129e02} & Vilafranca              & 0.9708 & 1132 \\
\textit{c03m093e04} & La Pobla de Benifassa   & 0.9653 & 1231 \\
\textit{c02m139e04} & Vistabella del Maestrat & 0.9552 & 1249 \\
\textit{c02m051e03} & Culla                   & 0.9391 & 1058 \\
\textit{c01m080e01} & Morella                 & 0.7663 &  980 \\
\bottomrule
\end{tabular}
\label{tab:cluster4_sensors}
\end{table}

\section{Model Architectures}
\label{sec:algo}

\subsection{Multilayer Perceptron}
A MLP is a class of feedforward neural networks designed to capture complex non-linear relationships within data through fully connected layers \cite{rumelhart1986learning}. The model consists of an input layer that receives the feature representation, followed by one or more hidden layers in which non-linear transformations are applied, and a final output layer that produces the network output. In an MLP, each neuron in a given layer is fully connected to all neurons in the subsequent layer, enabling the network to learn rich feature interactions through weighted connections and non-linear activation functions.

\subsection{Long Short-Term Memory}

LSTM networks are a specialised variant of RNNs designed to model long-term dependencies in sequential data \cite{hochreiter1997long}. An LSTM cell uses three gating mechanisms: input, forget, and output gates, to control how information is updated, stored, and passed forward through time. By regulating the flow of information at each time step, LSTM networks mitigate the vanishing gradient problem associated with standard RNNs and are particularly well-suited for time-series modelling tasks.

\subsection{Bidirectional Long Short-Term Memory}
BiLSTM networks extend standard LSTM architectures by processing sequential data in both forward and backward temporal directions \cite{schuster1997bidirectional}. This bidirectional processing enables the model to exploit past and future contextual information simultaneously at each time step, thereby enhancing its ability to capture complex temporal dependencies. BiLSTMs are particularly effective for time-series modelling tasks in which information from both preceding and subsequent observations contributes to accurate representation and learning.

\section{Experiments and Results}
\label{sec:exp_res}

\subsection{Feature Engineering and Data Scaling} 

To enhance the anomaly detection capabilities of the DL models, a set of engineered features is derived from the timestamp and temperature data. Raw timestamps are transformed into cyclical representations using sine and cosine functions (hour\_sin, hour\_cos, dow\_sin, dow\_cos) to preserve the periodic nature of daily and weekly cycles, ensuring that temporally adjacent values remain close in feature space. First- and second-order temporal derivatives are computed to capture the rate of temperature change and its acceleration, with an energy feature, defined as the squared first-order derivative, is included to emphasise periods of rapid temperature variation. In addition, statistical features based  on rolling windows, such as the mean, standard deviation, and range, are computed to provide local temporal context and capture short-term trends.

The engineered feature values are normalised prior to being fed into the neural network. StandardScaler is used for feature normalisation, which transforms the data to have zero mean and unit variance, as defined in Equation \ref{eq:standardscaler}

\begin{equation}
X_{scaled} = \frac{X - \mu}{\sigma}
\label{eq:standardscaler}
\end{equation}

Here, $X$ represents an individual data point in the time series. $\mu$ and $\sigma$ denote the mean and standard deviation of the training data, respectively. The scaler is fitted exclusively on the training data of each representative station, and the learned transformation parameters are applied to the corresponding test data, thereby ensuring that no information from the test period leaks into the training process.

\subsection{Training and Testing Data}

The temporal data for each station are partitioned into training and testing set using an 80/20 chronological split to preserve the sequential nature of the time-series data. The training set consists exclusively of normal temperature observations, ensuring that the autoencoder models learn to reconstruct only normal temperature behaviour. Due to the absence of comprehensive ground-truth labels for meteorological sensor anomalies, anomalies are synthetically simulated in the test set, drawing on real anomalous patterns identified in selected AVAMET stations during exploratory data analysis. The anomaly simulation procedure injects collective synthetic anomalies into otherwise normal temperature time-series while preserving the underlying temporal structure. The resulting anomalous sequences are labelled as 1, whereas normal temperature patterns are labelled as 0, enabling binary classification-based evaluation during model testing.

\subsection{Hyperparameter Search}

To identify the optimal set of hyperparameters, Bayesian optimisation was employed. This approach uses a probabilistic surrogate model to systematically explore the hyperparameter space, selecting promising configurations based on prior evaluations. To improve robustness and mitigate the risk of bias and overfitting, an expanding-window cross-validation strategy with five folds was adopted. 

\begin{table*}[htbp]
\centering
\caption{Hyperparameter search space used for Bayesian optimisation.}
\label{tab:hpo_search_space}
\begin{tabular}{llll}
\toprule
\textbf{Hyperparameter} & \textbf{BiLSTM} & \textbf{LSTM} & \textbf{MLP} \\
\midrule
\textit{hidden\_units} / \textit{lstm\_units} & [64, 256] & [64, 256] & [64, 256] \\
\textit{encoding\_dim} & [16, 128] & [16, 128] & [16, 96] \\
\textit{learning\_rate} & [10$^{-4}$, 10$^{-2}$] & [10$^{-4}$, 10$^{-2}$] & [10$^{-4}$, 10$^{-2}$] \\
\textit{dropout\_rate} & [0.1, 0.4] & [0.1, 0.4] & [0.0, 0.3] \\
\textit{batch\_size} & \{32, 64\} & \{32, 64\} & \{32, 64\} \\
\bottomrule
\end{tabular}
\end{table*}

Traditional k-fold cross-validation based on random shuffling is not suitable for time-series data, as it violates the chronological ordering of temporally dependent observations. In contrast, the expanding window cross-validation method partitions the dataset into sequential folds following the temporal order. The approach starts with an initial training window and progressively expands the training set by incorporating new observations in each subsequent fold, while validation is performed on the immediately following temporal segment. This strategy ensures that models are consistently evaluated on future unseen data, thereby closely mimicking real-world deployment conditions.

The performance of each hyperparameter configuration was evaluated using the average validation loss across the five folds. The hyperparameters optimised for each architecture, as summarised in Table \ref{tab:hpo_search_space}, varied slightly across models but generally included the number of hidden units, the dimensionality of the latent representation (\textit{encoding\_dim}), the learning rate (\textit{learning\_rate}) of the Adam optimiser, the dropout rate (\textit{dropout\_rate}), and the batch size (\textit{batch\_size}). A key aspect of this methodology is the use of \texttt{EarlyStopping} and \texttt{ReduceLROnPlateau} callbacks to mitigate overfitting and promote stable convergence. The hyperparameter set yielding the lowest average validation loss across the five folds was selected as the optimal configuration for each architecture, and all subsequent anomaly detection experiments were conducted using these configurations.

In addition to the evaluation metrics, the computational cost associated with Bayesian hyperparameter optimisation is also analysed across the evaluated models. MLP-based architectures exhibit substantially lower training times compared with RNN–based models. LSTM models show moderate computational requirements, whereas BiLSTM models incur the highest training cost due to their bidirectional processing structure. All models are trained and optimised using Bayesian hyperparameter optimisation on an NVIDIA GeForce RTX 5090 GPU (compute capability 12.0). The implementations are developed in Python using the TensorFlow and Keras frameworks. To support reproducibility and transparency, all source code and experimental results are publicly available \cite{SaadoonHammad2026}.


\subsection{Final Model Configuration}

The results of the hyperparameter optimisation are summarised in Table \ref{tab:optimal_hyperparameters} for the MLP, LSTM and BiLSTM autoencoder architectures. The selected configurations exhibit largely consistent parameter settings across clusters, with only minor variations observed in the latent dimensionality, batch size, and learning rate. These optimised hyperparameters are adopted for the final model training phase and subsequently evaluated on the test data to assess the anomaly detection performance of each algorithm.

\begin{table}[htbp]
\centering
\caption{Optimal hyperparameters obtained through Bayesian optimisation for each cluster and model architecture.}
\label{tab:optimal_hyperparameters}
\begin{tabular}{lcccc}
\toprule
\textbf{Hyperparameters} & \multicolumn{4}{c}{\textbf{Clusters}} \\
\cmidrule(lr){2-5}
 & \textbf{C1} & \textbf{C2} & \textbf{C3} & \textbf{C4} \\
\midrule
\multicolumn{5}{l}{\textit{\textbf{LSTM Autoencoder}}} \\
\midrule
LSTM Units & 256 & 254 & 256 & 256 \\
Encoding Dimension & 64 & 60 & 64 & 64 \\
Learning Rate & 0.001 & 0.00087 & 0.001 & 0.001 \\
Dropout Rate & 0.10 & 0.103 & 0.10 & 0.10 \\
Batch Size & 32 & 32 & 32 & 32 \\
\midrule
\multicolumn{5}{l}{\textit{\textbf{BiLSTM Autoencoder}}} \\
\midrule
LSTM Units & 256 & 256 & 238 & 256 \\
Encoding Dimension & 92 & 128 & 124 & 95 \\
Learning Rate & 0.001 & 0.001 & 0.0009 & 0.001 \\
Dropout Rate & 0.1 & 0.1 & 0.1 & 0.1 \\
Batch Size & 32 & 32 & 16 & 64 \\
\midrule
\multicolumn{5}{l}{\textit{\textbf{MLP Autoencoder}}} \\
\midrule
Hidden Units & 256 & 256 & 256 & 256 \\
Encoding Dimension & 96 & 96 & 80 & 96 \\
Learning Rate & 0.0001 & 0.000250 & 0.0001 & 0.000242 \\
Dropout Rate & 0.0 & 0.0 & 0.0 & 0.0 \\
Batch Size & 32 & 64 & 32 & 64 \\
\bottomrule
\end{tabular}
\small
\end{table}

\subsection{Evaluation Metrics}

The anomaly detection performance is evaluated using standard binary classification metrics. Let TP (True Positive), TN (True Negative), FP (False Positive), and FN (False Negative) denote the elements of the confusion matrix:

\begin{equation}
\text{Precision} = \frac{TP}{TP + FP}
\end{equation}

\begin{equation}
\text{Recall} = \frac{TP}{TP + FN}
\end{equation}

\begin{equation}
\text{F1-Score} = 2 \times \frac{\text{Precision} \times \text{Recall}}{\text{Precision} + \text{Recall}}
\end{equation}

Precision measures the proportion of correctly identified anomalies among all predicted anomalies, while Recall quantifies the proportion of actual anomalies that are successfully detected. The F1-Score, defined as the harmonic mean of Precision and Recall, provides a balanced performance measure, particularly when both false positives and false negatives are relevant. In addition, the Area Under the Precision-Recall Curve (PR-AUC) is used as a threshold-independent metric, which is especially appropriate for imbalanced datasets, where anomalous events occur infrequently.

\section{Results}
\label{sec:res}

This section presents the results of our proposed method. The generalisability of the models is evaluated both within individual communities and across different communities, in order to assess how well models trained on a representative station transfer to other stations with similar or different characteristics. 

\subsection{Performance of Cluster 1}

Table \ref{tab:results_c1} summarises the generalisation performance of the different autoencoder architectures trained using the representative station \textit{c02m042e02} from Cluster 1. When evaluated on stations belonging to the same cluster, the MLP autoencoder exhibits consistently strong performance, with precision values ranging from 0.8740 to 0.9176. However, when applied to stations from other clusters, a noticeable decrease in precision and PR-AUC is observed, indicating an increase in false positives detections. 

The LSTM-based model achieves high precision values when evaluated within its own cluster, reaching a maximum precision value of 0.9197. Across clusters, its performance remains largely stable for some communities, while greater variability is observed for others, where precision decreases due to an increase in false positives.

Similarly, the BiLSTM autoencoder demonstrates strong performance across all stations within Cluster 1, with precision values ranging between 0.8243 to 0.9141. For cross-cluster evaluation, the BiLSTM model achieves high precision values when applied to stations from Clusters C3 and C4. In contrast, a degradation in precision is observed for Cluster C2, where the model exhibits an increased number of false positives, resulting in a precision value of 0.5811. Overall, the BiLSTM architecture exhibits robust within-cluster performance and competitive cross-cluster generalisation.

In summary, all evaluated models demonstrate strong performance and generalisability within the cluster and across communities. However, reduced performance is observed when models are applied to Cluster C2, with lower precision values particularly evident for the LSTM and BiLSTM architectures.

\begin{table*}[htbp]
\centering
\caption{Cluster 1 model generalisability: performance of autoencoder models trained on the representative station and evaluated on intra- and inter-cluster stations.}
\label{tab:results_c1}

\begin{tabular}{lllcccc}
\toprule
\textbf{Method} & \textbf{Train Station} & \textbf{Test Station} & \textbf{Precision} & \textbf{Recall} & \textbf{F1-Score} & \textbf{PR-AUC} \\
\midrule

 \multirow{6}{*}{MLP} 
    & \multirow{6}{*}{c02m042e02} 
    & c02m042e02 (C1) & 0.9047 & 0.9959 & 0.9481 & 0.7932 \\
 &  & c02m014e04 (C1) & 0.9176 & 0.9740 & 0.9450 & 0.8742 \\
 &  & c04m072e01 (C1) & 0.8740 & 0.9926 & 0.9295 & 0.7277 \\
&  & c06m136e02 (C2) & 0.7817 & 0.9939 & 0.8751 & 0.7250 \\
 &  & c08m046e01 (C3) & 0.7973 & 0.9639 & 0.8727 & 0.7676 \\
 &  & c01m129e02 (C4) & 0.8832 & 0.9949 & 0.9358 & 0.7234 \\

\cmidrule{3-7}

 \multirow{6}{*}{LSTM} 
& \multirow{6}{*}{c02m042e02} 
    & c02m042e02 (C1) & 0.8797 & 0.9959 & 0.9342 & 0.7857 \\
 &  & c02m014e04 (C1) & 0.9197 & 0.9699 & 0.9441 & 0.8637 \\
 &  & c04m072e01 (C1) & 0.8574 & 0.9944 & 0.9208 & 0.7186 \\
 &  & c06m136e02 (C2) & 0.6764 & 0.9939 & 0.8050 & 0.6674 \\
 &  & c08m046e01 (C3)& 0.8665 & 0.9518 & 0.9072 & 0.7753 \\
 &  & c01m129e02 (C4) & 0.8832 & 0.9949 & 0.9358 & 0.7212 \\
\cmidrule{3-7}
 \multirow{6}{*}{BiLSTM} 
& \multirow{6}{*}{c02m042e02} 
 & c02m042e02  (C1)  & 0.8702 & 0.9959 & 0.9288 & 0.7502 \\
 &  & c02m014e04  (C1)  & 0.9141 & 0.9740 & 0.9431 & 0.8552 \\
 &  & c04m072e01  (C1) & 0.8243 & 0.9944 & 0.9014 & 0.7169 \\
 &  & c06m136e02  (C2) & 0.5811 & 0.9939 & 0.7334 & 0.6682 \\
&  & c08m046e01  (C3) & 0.8551 & 0.9598 & 0.9044 & 0.7755 \\
&  & c01m129e02  (C4) & 0.8832 & 0.9949 & 0.9358 & 0.7220 \\
\bottomrule
\end{tabular}%

\end{table*}

\subsection{Performance of Cluster 2}
Table \ref{tab:results_c2} summarises the generalisability of the various autoencoder algorithms trained using the most representative sensor from C2 with id \textit{c06m136e02}. The precision shows substantial variability within the cluster, ranging from 0.2641 to 0.6864, and the values are notably lower than those observed for Cluster 1. In cross-cluster evaluation, the model exhibits reduced performance on C3, with a precision value of 0.1227, accompanied by a decrease in PR-AUC, indicating an increased number of false positives.

The LSTM model follows a similar trend, with precision values of 0.3748 and 0.5703 for within-cluster evaluations. Across clusters, the model shows a marked reduction in performance on C3 with precision value of 0.1785, while more stable performance is observed for C1 and C4. For BiLSTM, the performance degrades further relative to the MLP and LSTM models, both within the cluster and  across clusters, with the lowest precision value of 0.0694 reported for station id \textit{c08m046e01} from C3. These results indicate an elevated false positive rates, meaning that the models tend to incorrectly flag normal temperature patterns as anomalous. 

Overall, for all algorithms trained on data from the representative station of C2, the performance suggests limited generalisability, both within the cluster and across other communities.

\begin{table*}[htbp]
\centering
\caption{Cluster 2 model generalisability: performance of autoencoder models trained on the representative station and evaluated on intra- and inter-cluster stations.}
\label{tab:results_c2}

\begin{tabular}{lllcccc}
\toprule
 \textbf{Method} & \textbf{Train Station} & \textbf{Test Station} &  \textbf{Precision} & \textbf{Recall} & \textbf{F1-Score} & \textbf{PR-AUC}  \\
\midrule
 \multirow{6}{*}{MLP} 
    & \multirow{6}{*}{c06m136e02} & c06m136e02 (C2) & 0.4453 & 0.9959 & 0.6154 & 0.6530 \\
 &  & c05m040e13 (C2) & 0.6864 & 0.9955 & 0.8125 & 0.7929 \\
 &  & c05m031e06 (C2) & 0.2641 & 0.9982 & 0.4177 & 0.5410 \\
 &  & c02m042e02 (C1) & 0.6452 & 0.9959 & 0.7831 & 0.7893 \\
 &  & c08m046e01 (C3) & 0.1227 & 0.9980 & 0.2186 & 0.4962 \\
 &  & c01m129e02 (C4) & 0.6637 & 0.9949 & 0.7962 & 0.7220 \\
\cmidrule{3-7}
 \multirow{6}{*}{LSTM} 
& \multirow{6}{*}{c06m136e02} 
& c06m136e02 (C2) & 0.3748 & 0.9959 & 0.5446 & 0.6253 \\
 &  & c05m040e13 (C2) & 0.5703 & 0.9955 & 0.7252 & 0.7852 \\
 &  & c05m031e06 (C2) & 0.3697 & 0.9876 & 0.5380 & 0.5702 \\
 &  & c02m042e02 (C1) & 0.6845 & 0.9959 & 0.8113 & 0.7562 \\
&  & c08m046e01 (C3) & 0.1785 & 0.9779 & 0.3019 & 0.6026 \\
 &  & c01m129e02 (C4) & 0.6277 & 0.9949 & 0.7697 & 0.7118 \\
\cmidrule{3-7}
\multirow{6}{*}{BiLSTM} 
& \multirow{6}{*}{c06m136e02} & c06m136e02 (C2) & 0.2594 & 0.9959 & 0.4116 & 0.6198 \\
 &  & c05m040e13 (C2) & 0.4018 & 0.9955 & 0.5725 & 0.7615 \\
 &  & c05m031e06 (C2) & 0.1192 & 1.0000 & 0.2130 & 0.4233 \\
 &  & c02m042e02 (C1)& 0.3031 & 0.9959 & 0.4647 & 0.7039 \\
 &  & c08m046e01 (C3) & 0.0694 & 0.9980 & 0.1297 & 0.4922 \\
 &  & c01m129e02 (C4) & 0.2739 & 0.9949 & 0.4296 & 0.6768 \\

\bottomrule
\end{tabular}%

\end{table*}

\subsection{Performance of Cluster 3}

The performance of models trained on the representative station \textit{c08m046e01} from C3 is summarised in the Table \ref{tab:results_c3}. When the MLP-model is evaluated on stations belonging to the same cluster, it exhibits strong performance, with precision values ranging from 0.8582 to 0.8679, indicating reliable anomaly discrimination within the cluster. However, when evaluated on stations from other clusters, a noticeable reduction in precision and PR-AUC is observed for Cluster 2. 

The LSTM-based model follows a similar behaviour, achieving high precision on the representative station and maintaining stable performance across other stations within Cluster 3, while cross-cluster evaluation reveals reduced precision for Cluster 2 and moderate degradation for Cluster 1 and 4. The BiLSTM model demonstrates comparable intra-cluster performance, indicating effective anomaly detection with a low false positive rate. In cross-cluster scenarios, the BiLSTM maintains reasonable performance for Cluster 1 and 4,  exhibits a marked reduction in precision when applied to Cluster 2, suggesting  limited transferability of the learned representations to this community.

Overall, the results indicate that while all models generalise well within Cluster 3 and, to some extent, across other communities. Nevertheless, Cluster 2 consistently remains the most challenging case, with performance degradation consistently observed across all architectures.

\begin{table*}[htbp]
\centering
\caption{Cluster 3 model generalisability: performance of autoencoder models trained on the representative station and evaluated on intra- and inter-cluster stations.}
\label{tab:results_c3}

\begin{tabular}{lllcccc}
\toprule
\textbf{Method} & \textbf{Train Station} & \textbf{Test Station} &  \textbf{Precision} & \textbf{Recall} & \textbf{F1-Score} & \textbf{PR-AUC}  \\
\midrule
 \multirow{6}{*}{MLP} 
    & \multirow{6}{*}{c08m046e01} 
    & c08m046e01 (C3) & 0.8679 & 0.9237 & 0.8949 & 0.7721 \\
 &  & c08m017e01 (C3) & 0.8582 & 0.8110 & 0.8339 & 0.7084 \\
 &  & c03m096e01 (C3) & 0.8603 & 0.7270 & 0.7880 & 0.7958 \\
 &  & c02m042e02 (C1) & 0.8846 & 0.8038 & 0.8423 & 0.7886 \\
 &  & c06m136e02 (C2) & 0.7785 & 0.9898 & 0.8715 & 0.6709 \\
 &  & c01m129e02 (C4) & 0.8424 & 0.7032 & 0.7665 & 0.7201 \\
\cmidrule{3-7}
 \multirow{6}{*}{LSTM} 
& \multirow{6}{*}{c08m046e01} 
& c08m046e01 (C3) & 0.8547 & 0.8976 & 0.8756 & 0.7656 \\
 &  & c08m017e01 (C3) & 0.8426 & 0.8557 & 0.8491 & 0.6944 \\
 &  & c03m096e01 (C3) & 0.8477 & 0.6656 & 0.7457 & 0.7923 \\
 &  & c02m042e02 (C1) & 0.8537 & 0.8106 & 0.8316 & 0.7518 \\
 &  & c06m136e02 (C2) & 0.5596 & 0.9000 & 0.6901 & 0.6097 \\
 &  & c01m129e02 (C4) & 0.8597 & 0.8061 & 0.8320 & 0.7216 \\
\cmidrule{3-7}
 \multirow{6}{*}{BiLSTM} 
& \multirow{6}{*}{c08m046e01}
& c08m046e01 (C3)& 0.8528 & 0.8956 & 0.8737 & 0.7486 \\
 &  & c08m017e01 (C3) & 0.8275 & 0.8076 & 0.8174 & 0.6835 \\
 &  & c03m096e01 (C3) & 0.8446 & 0.6503 & 0.7348 & 0.7835 \\
 &  & c02m042e02 (C1) & 0.8104 & 0.7221 & 0.7637 & 0.7163 \\
 &  & c06m136e02 (C2) & 0.5730 & 0.9531 & 0.7157 & 0.5879 \\
 &  & c01m129e02 (C4) & 0.8016 & 0.6610 & 0.7246 & 0.6599  \\

\bottomrule
\end{tabular}%

\end{table*}

\subsection{Performance of Cluster 4}

The generalisability performance of models trained on the representative station \textit{c01m129e02} from Cluster 4 is summarised in Table \ref{tab:results_c4}. The intra-cluster performance of MLP-based model exhibits consistently high precision values, remaining  stable across all stations within the clusters, with values ranging from 0.8662 to 0.8877. When evaluated  on stations from other clusters, the model maintains high precision for Cluster 1, in some cases exceeding the performance observed within its own cluster. In contrast, reduced performance is observed for Clusters 2 and 3, where precision and PR-AUC decrease to 0.7424 and 0.7031, respectively.

The LSTM-based model demonstrates similar pattern, showing stable performance within Cluster 4.  Inter-cluster evaluation, however, the model exhibits a marked reduction in precision, reaching 0.6134 when applied to other communities. Likewise, the BiLSTM architecture follows a comparable trend, achieving strong intra-cluster performance while experiencing reduced precision in cross-cluster scenarios, particularly when evaluated on Cluster 2.

Overall, the results indicate that all evaluated models generalise effectively within Cluster 4, reflecting robust intra-cluster performance. Across communities, performance varies depending on the target cluster, with consistent degradation observed when models are applied to Cluster 2, suggesting limited transferability of models trained on Cluster 4 to this community.

\begin{table*}[htbp]
\centering
\caption{Cluster~4 model generalisability: performance of autoencoder models trained on the representative station and evaluated on intra- and inter-cluster stations.}
\label{tab:results_c4}

\begin{tabular}{lllcccc}
\toprule
\textbf{Method} & \textbf{Train Station} & \textbf{Test Station} &  \textbf{Precision} & \textbf{Recall} & \textbf{F1-Score} & \textbf{PR-AUC}  \\
\midrule
 \multirow{6}{*}{MLP} 
& \multirow{6}{*}{c01m129e02} & c01m129e02 (C4) & 0.8832 & 0.9949 & 0.9358 & 0.7236 \\
 &  & c03m093e04 (C4) & 0.8877 & 0.9948 & 0.9382 & 0.7878 \\
 &  & c01m080e01 (C4) & 0.8662 & 0.9939 & 0.9257 & 0.7902 \\
 &  & c02m042e02 (C1) & 0.9047 & 0.9959 & 0.9481 & 0.7929 \\
 &  & c06m136e02 (C2) & 0.7424 & 0.9939 & 0.8499 & 0.7194 \\
 &  & c08m046e01 (C3) & 0.7031 & 0.9558 & 0.8102 & 0.7480 \\
\cmidrule{3-7}
 \multirow{6}{*}{LSTM} 
& \multirow{6}{*}{c01m129e02} 
& c01m129e02 (C4)  & 0.8806 & 0.9949 & 0.9343 & 0.7218 \\
 &  & c03m093e04 (C4) & 0.8826 & 0.9948 & 0.9353 & 0.7786 \\
 &  & c01m080e01 (C4) & 0.8647 & 0.9939 & 0.9248 & 0.7591 \\
 &  & c02m042e02 (C1) & 0.8776 & 0.9959 & 0.9330 & 0.7472 \\
 &  & c06m136e02 (C2) & 0.6134 & 0.9939 & 0.7586 & 0.6713 \\
 &  & c08m046e01 (C3) & 0.8148 & 0.9538 & 0.8788 & 0.7631 \\
\cmidrule{3-7}
 \multirow{6}{*}{BiLSTM} 
& \multirow{6}{*}{c01m129e02} & c01m129e02 (C4) & 0.8832 & 0.9949 & 0.9358 & 0.7213 \\
 &  & c03m093e04 (C4) & 0.8856 & 0.9843 & 0.9323 & 0.7618 \\
 &  & c01m080e01 (C4) & 0.8677 & 0.9939 & 0.9266 & 0.7510 \\
 &  & c02m042e02 (C1) & 0.8926 & 0.9959 & 0.9414 & 0.7693 \\
 &  & c06m136e02 (C2) & 0.7099 & 0.9939 & 0.8282 & 0.6251 \\
 &  & c08m046e01 (C3) & 0.8616 & 0.9498 & 0.9035 & 0.7489 \\
\bottomrule
\end{tabular}%

\end{table*}

\section{Conclusion}
\label{sec:conclusion}

This work presents a comprehensive framework for anomaly detection in IoT temperature sensor networks, leveraging the CoI paradigm in combination with deep autoencoder architectures. By exploiting temporal correlations, spatial proximity, and elevation similarities, the proposed approach employs hierarchical clustering to organise stations into communities, enabling efficient model sharing within sensor groups. This community-based approach addresses the limitations of one-model-per-device strategy, which is often impractical for IoT networks where individual devices may lack sufficient data or computational resources to train their own models. It also facilitates the evaluation of model generalisability across stations with similar characteristics. 

The experimental evaluation conducted on 41 meteorological stations in the province of Castelló demonstrates that community-based model sharing for anomaly detection can achieve strong performance when stations exhibit coherent temporal and spatial behaviour. Models trained on representative stations from Clusters 1, 3 and 4 achieved high anomaly detection performance, with precision ranging between 0.8275 to 0.9176, indicating the ability of the models to effectively capture normal temperature patterns and distinguish anomalous behaviour.

The cross-cluster generalisability analysis further shows promising transferability for models associated with Clusters 1, 3 and 4, with precision values remaining above 0.80 when evaluated on stations from other communities. In contrast, reduced generalisability is observed for Cluster 2. Models trained on representative stations from other clusters exhibit notable performance degradation when applied to Cluster 2, with precision values decreasing to 0.5811. Similarly, models trained on the representative station of Cluster 2 demonstrate limited transferability when evaluated on stations from other communities, with precision values dropping 0.0694 in the most challenging cases.

In summary, the proposed framework demonstrates that community-based model sharing is a viable strategy for IoT sensors networks characterised by strong temporal and spatial coherence. By shifting from a one-model-per-sensor paradigm to a one-per-community, the framework has the potential to significantly reduce computational requirements, lower operational costs, and improve scalability in large-scale IoT deployments.

As future work, we plan to explore 
the deployment of the trained models on resource-constrained edge devices, such as microcontrollers, by converting them into lightweight models optimised for embedded and edge computing environments. We also aim to develop a region-based model distribution framework, where each geographical community has a dedicated fog device. The trained models will be stored in a decentralized repository, enabling automatic assignment of the appropriate community-specific model to each fog/edge device for real-time anomaly detection on stations within that region. In addition, we aim to explore the interpretability of the proposed models by incorporating eXplainable Artificial Intelligence (XAI) techniques to better understand which features—temporal, spatial, or derived—most strongly influence anomaly detection decisions within each community. Such analysis is expected to provide deeper insights into the factors contributing to the reduced performance observed for Cluster 2.

\section{Acknowledgements}
The authors thank the Associació Valenciana de Meteorologia `Josep Peinado' (AVAMET) for providing the data from its stations used in this research. This publication is part of the project PID2022-141813OB-I00 funded by MCIN\slash AEI\slash10.13039\slash501100011033 and by ERDF/EU. Project supported by a 2024 Leonardo Grant for Scientific Research and Cultural Creation from the BBVA Foundation.

\bibliographystyle{unsrtnat}
\footnotesize
\bibliography{references}

\end{document}